\documentclass{article}
\usepackage{spconf}
\usepackage{amsmath}
\usepackage{graphicx}

\usepackage{hyperref}       
\usepackage{url}            
\usepackage{amssymb}

\newcommand{\datasetName}{VisionScores}
\newcommand{\datasetURL}{\url{https://github.com/alroamz/VisionScores}}

\title{\MakeUppercase{\datasetName{}} - A SYSTEM-SEGMENTED IMAGE SCORE DATASET FOR DEEP LEARNING TASKS}
\name{Alejandro Romero Amezcua and Mariano José Juan Rivera Meraz}
\address{Centro de Investigación en Matemáticas A.C.\\
	Apdo. Postal 36023, Guanajuato, Gto., México}
\begin{document}
\ninept
\maketitle
%
\begin{abstract}
\datasetName{} presents a novel proposal being the first system-segmented image score dataset, aiming to offer structure-rich, high information-density images for machine and deep learning tasks.
Delimited to two-handed piano pieces, it was built to consider not only certain graphic similarity but also composition patterns, as this creative process is highly instrument-dependent.
It provides two scenarios in relation to composer and composition type. 
The first, formed by 14k samples, considers works from different authors but the same composition type, specifically, Sonatinas.
The latter, consisting of 10.8K samples, presents the opposite case, various composition types from the same author, being the one selected Franz Liszt. 
All of the 24.8k samples are formatted as grayscale jpg images of $128 \times 512$ pixels. 
\datasetName{} supplies the users not only the formatted samples but the systems' order and pieces' metadata. Moreover, unsegmented full-page scores and the pre-formatted images are included for further analysis.\\
\datasetName{} is freely available in \datasetURL{}.
\end{abstract}
%
\section{Introduction}
\label{sec:intro}
The rapid advancements in generative artificial intelligence (GenAI) have revolutionized content creation across various domains, including text, images, audio, and video. This has led to outstanding applications in fields as diverse as creative, industry, scientific research, education, and more. GenAI not only enhances productivity but also redefines traditional creative processes, making it a cornerstone of modern computational research.\\
Within this broader context, symbolic music generation has emerged as a significant area of exploration. Symbolic music refers to the representation of musical compositions in structured formats such as music sheets or MIDI files, enabling both human-readable and machine-processable interpretations. Researchers have developed a variety of machine learning models that exploit state-of-the-art architectures. For instance \textit{Music Transformer} \cite{music_transformer} employs the \textit{Transformer} architecture for harmonization tasks, while \textit{MusicBERT} \cite{musicbert} and \textit{MRBERT} \cite{mrbert}, based on \textit{RoBERTa} \cite{roberta} and \textit{BART} \cite{bart} respectively, extend natural language processing methodologies to tasks such as melody completion, accompaniment generation, and genre classification.\\
The success of these models has been heavily dependent on the availability of high-quality datasets. Symbolic music datasets are predominantly built around \textit{MIDI} and piano roll, due to the abundance of works in these formats. Although there are several tools for converting musical notation to \textit{MIDI} encoding, human intervention is often required for corrections. Additionally, this codification has some representation issues compared with music sheets, for instance, precise note duration, annotations, ornaments, among others. This led to the consideration of using musical scores in image format. While several score datasets in image format already exist, they are predominantly designed for \textit{Optical Music Recognition (OMR)} \cite{omr}, which limits their applicability to broader tasks. Their narrow focus often render them unsuitable for more general applications in symbolic music processing for machine learning.
Therefore, a crucial step is the construction of dedicated datasets, which fulfill the characteristics for supporting a wide range of applications. \\
The \datasetName{} dataset introduces a novel and highly structured collection of system-segmented image scores tailored for machine learning applications in symbolic music processing. \datasetName{} is designed to support a broader range of tasks by offering 24,810 standardized grayscale images derived from two-handed piano compositions. It comprises two curated scenarios—Sonatinas by multiple composers and varied works by Franz Liszt—ensuring both structural consistency and stylistic diversity. Each sample includes detailed metadata and maintains system-level order, enabling the exploration of sequential patterns and hierarchical relationships within musical compositions.
A detailed description of the construction process, as well as the challenges of each of its phases, is presented: construction approach (section \ref{sec:dataset}), data collection (section \ref{sec:data_collection}), image segmentation (section \ref{sec:segmentation}), results of the application of developed methods (section \ref{sec:results}) and the overall constructed dataset (section \ref{sec:conclusions}).



\section{Related work and motivation}
\label{sec:related}
As mentioned in the previous section, there are already score datasets in image format; nevertheless, they are focused on Optical Music Recognition (OMR) which involves tasks such as automatic conversion of images of music scores into machine-readable formats, symbol classification, and so forth. This makes these datasets less suitable or even unusable for other tasks.
Despite this fact, it is worth performing a general review of these datasets, as they are, or were, considered a reference in the state of the art.

\begin{itemize}
\item \textit{CVC-MUSCIMA} \cite{cvc_muscima}: Developed by the \textit{Computer Vision Center (CVC)} \cite{cvc}, was design for the \textbf{recognition of handwritten music scores}. It is a collection of 1,000 images created by 50 different musicians, designed for tasks such as writer identification and staff removal. It is widely used for benchmarking in OMR and provides a robust ground truth for evaluating algorithms in symbol detection and writer identification.
\item \textit{MUSCIMA++} \cite{muscima_pp}: Is an extended and enhanced version of the previous dataset. Contains 91,255 annotated symbols across 140 images. It provides detailed annotations in the form of bounding boxes, pixel masks, and relationships between symbols, forming the \textit{MUSCIMA++ Notation Graph (MuNG)}. This dataset is specifically designed for \textbf{musical symbol detection, classification, and notation reconstruction}, offering a bridge between low-level and high-level OMR tasks.
\item \textit{DeepScores} \cite{deepscores}: A collection of 300,000 high-quality full-page musical printed scores images, which include various genres, composition types, instruments, musical ensembles, etc. It was designed to research \textbf{small object recognition and scene understanding}. In the context of OMR, \textit{DeepScores} provides a rich and diverse set of musical symbols and notations, nearly 100 million, enabling the development of more accurate and robust systems for automatic music transcription and analysis. Its focus in small objects, allows for extending its relevance beyond optical music recognition (OMR) to broader computer vision applications. 

\item \textit{AudioLabs - Measure Bounding Box Annotation} \cite{audiolabsv1}: \textbf{Focused on measure detection and bounding box annotation of music sheets}. It contains 940 full-page score images with 24,329 annotated bounding boxes for measures, staves, and systems. As DeepScores, the music sheets are not of a particular type. The dataset is primarily used for training and evaluating models in measure recognition and layout analysis in music sheets.
\end{itemize}

\begin{figure}[t] 
    \begin{minipage}[b]{.48\linewidth}
      \centering
      \centerline{\includegraphics[width=4.0cm]{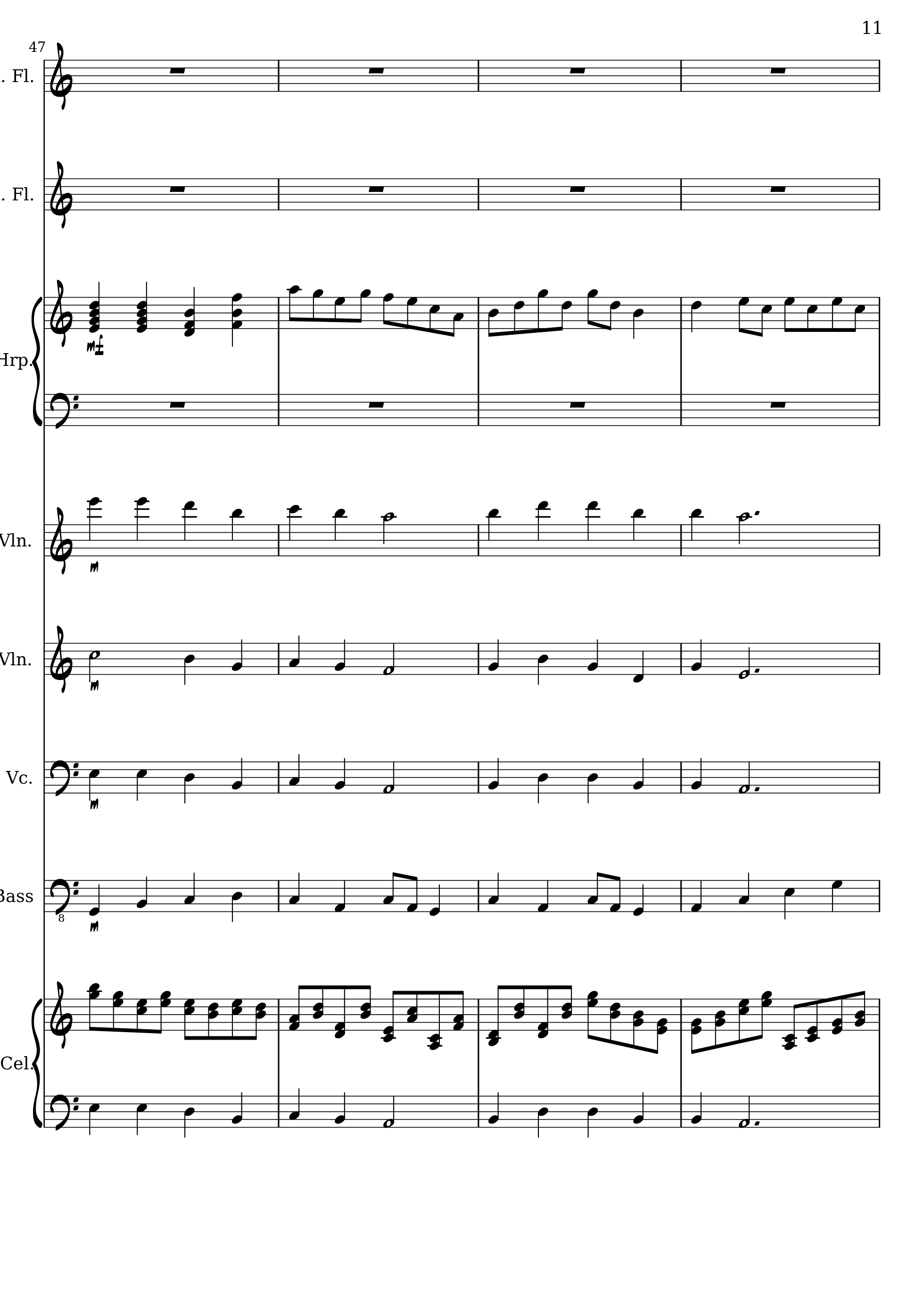}}
      \centerline{(a)}\medskip
    \end{minipage}
    \hfill
    \begin{minipage}[b]{0.48\linewidth}
      \centering
      \centerline{\includegraphics[width=4.0cm]{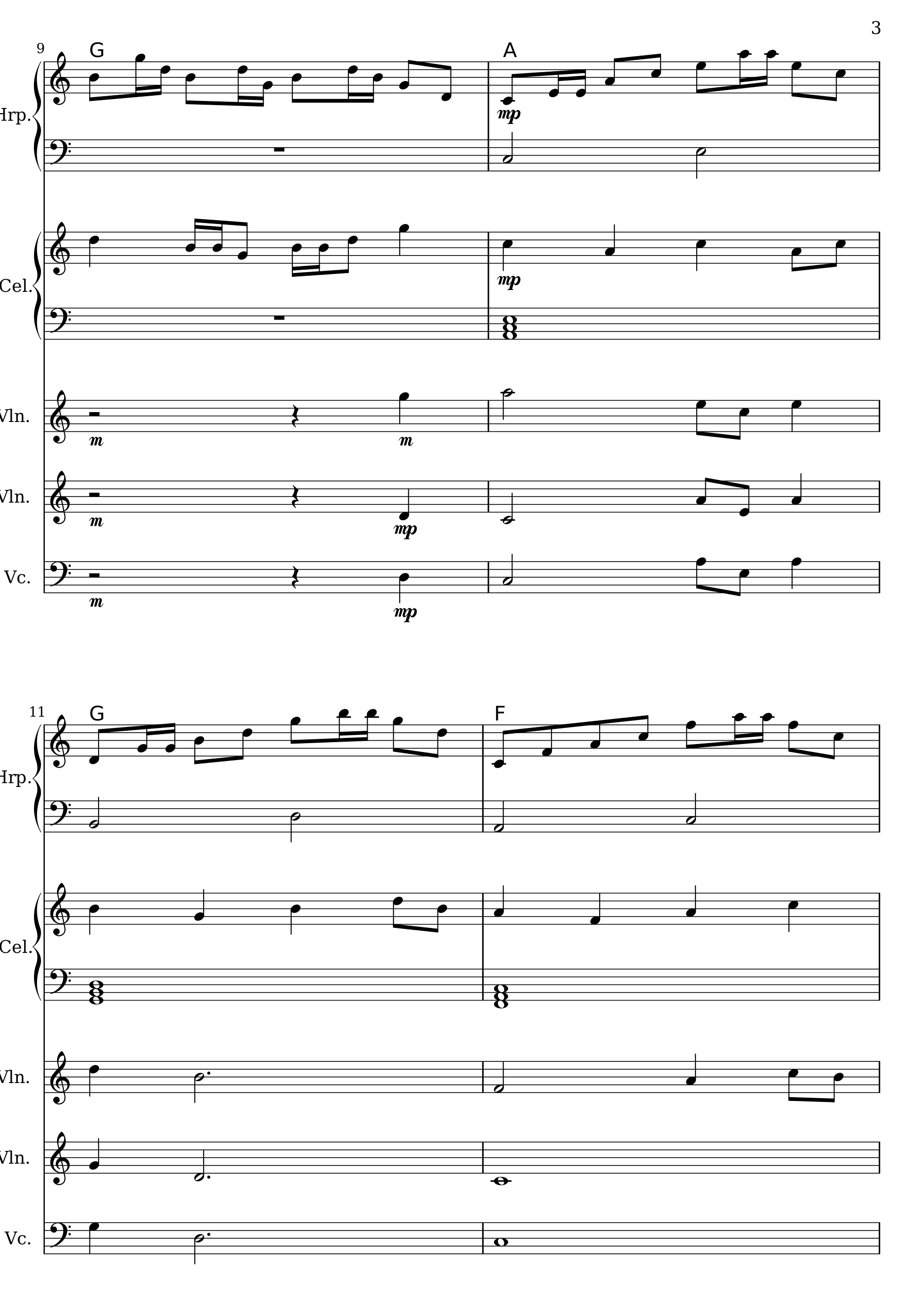}}
      \centerline{(b)}\medskip
    \end{minipage}
    \caption{Examples from the DeepScores dataset which shows the structural variety between samples. (a) A Single system for a 8-instrument ensemble. (b) 2 systems for a 5-instrument ensemble.}
    \label{fig:deepscores_example}
\end{figure}

It is important to acknowledge that there are additional music score datasets in image format. Nonetheless, the previous were considered as the most relevant in the literature and for this research. For overall inspection of these other available datasets, consult \cite{omr_datasets}.\\ 
Note that \textit{CVC-MUSCIMA} and \textit{MUSCIMA++} are formed by handwritten scores, a characteristic that limits their application, e.g. manuscript translation into machine-readable formats. On the other hand, \textit{DeepScores} has a broader application which may extend beyond OMR to general computer vision tasks, such as segmentation. Unfortunately, although full-page format could be difficult to manage, the wide range of musical ensembles, composition types, instruments, etc., limits the application of the dataset in content-dependent musical tasks, namely symbolic music generation. The same complication emerges in the \textit{AudioLabs - Measure Bounding Box Annotation} dataset.\\ 
In order to exhibit these differences, 2 samples of DeepScores are shown in figure \ref{fig:deepscores_example}. At first glance, any non-musician can identify the images as music scores. However, upon closer examination, distinct visual differences become apparent; for instance, one is divided into two sections, whereas the other remains undivided. In this particular example, the differences arise from the fact that the pages belong to pieces composed for different ensembles. Analogous to this, several dissimilarities can be enlisted between pieces that do not correspond to the same category. As a consequence of its intrinsic structural difference, the formatting of the data without negatively impacting its quality (image deformation, missing symbols, etc.) could be nearly impossible, which in turn may translate into a poor performance when using this formatted data as the training input for neural networks. Moreover, these models generally benefit from visual diversity, as long as the inherent structure of the processed data remains consistent; for example, images of dogs may vary by breed, yet all will share common features such as having four legs.\\


\section{\MakeUppercase{\datasetName{}} dataset}
\label{sec:dataset}
Taking into consideration the available datasets' characteristics, the need for a new dataset becomes evident. This necessity is the one that \textit{\datasetName{}} aims to resolve, to provide a dataset that fulfills the characteristics necessary for making it suitable as training data. These desirable features can be divided into 2 aspects: content and file. In relation to the first, structural similarity was needed. As for the image itself, the requirement for format uniformity without the loss of quality was imposed. These 2 conditions modeled the construction process of the dataset, which is discussed along this section.

\subsection{Content constraint} 
\label{sec:data_collection}
The initial step was to find a source from where the data could be collected. Since what was required were music sheets, \textit{IMSLP}\cite{imslp}
was the appropriate platform as it is a free access project dedicated to the collecting of this type of data, providing a wide variety of works of different genres, styles, authors, instruments, and so on. Most of the scores on the page are copyright free; however, certain pieces are not, which could potentially limit, or invalidate, the dataset to be distributed or even used. In order to cope with this drawback, a simple solution was followed: not to consider these works. Most pieces on IMSLP are available in various formats, but PDFs are the most common and were deemed suitable for dataset construction.\\
Once the data source was established, the next step was to define the specific type of musical scores to be used. Primarily, it was considered that the intrinsic properties of music ensembles composition would translate to generally sparse music sheets, as well as complicated visual variations depending on the specific ensemble. Due to this, the decision of considering works destined to be played on one instrument was taken. As for the selection of a particular instrument, it was considered that the piano was the most suitable option, as it is the one with more stand-alone pieces. Finally, to fully establish the fundamental characteristics of the data, two-handed pieces were selected within the variety of piano repertoire.\\ 
The delimitation of score types aimed to maintain structural consistency within the dataset. However, two-handed piano scores encompass various composition types. Further narrowing the selection to a specific type could provide valuable applications, as different pieces would likely share compositional similarities. Conversely, such a restriction might be overly limiting, reducing the dataset’s applicability. To balance these aspects, an alternative approach was added: restricting the dataset by composer. This resulted in two distinct scenarios. The first includes works from multiple composers but within the same composition type. Sonatinas were chosen due to their relatively short length and low musical complexity. The second scenario takes the opposite approach, featuring various composition types from a single composer. Franz Liszt was selected for two reasons: he had the largest amount of available data under the established criteria, and his works exhibit a high level of technical difficulty, contrasting with the simplicity of Sonatinas.\\
To obtain the data present on the page, the use of web scraping was necessary, interacting with the site by its HTML responses and collecting the information that was considered necessary. For each sheet, if available, the details about author, title and key were saved. The download link was used to obtain the PDF file, but it was not kept as a measure of security and respect to the page.
This decision was made to comply with IMSLP's terms and conditions. \\
Subsequently to the download, a process of manual selection of the files that were suitable for the dataset was carried out. The selection criterion was established according to the visual quality of the files: the closest to black and white images was the best, discarding manuscripts and scans in bad condition or sepia tones. This criterion was established due to the low color uniformity in the images at these tones, which made the process of converting them to black and white, with precise results very time-consuming. \\
Finally, once the PDF files were selected, they were converted to images, sheet-to-sheet. A manual selection process was required for deleting non-useful converted pages, such as covers, indexes, etc. This resulted in a set of two-handed piano music sheet images, composed of clusters of bars that extend to the width of the page called \textit{systems}. 


\subsection{Format constraint} 
\label{sec:segmentation}
Format uniformity without the loss of quality was the second condition, as \datasetName{} dataset's samples need to be shaped for its use in deep learning tasks while maintaining the integrity of its musical information, i.e. its symbology and structure. 
This could only be achieved by selecting the systems as the samples of the dataset, as full page scores are overly large and individual bars have a high variance in their length.\\
It is essential to recognize that individual systems cannot be classified by composition type or attributed to a specific composer, as these characteristics emerge from the work as a whole, i.e., from the ordered sequence of systems and the information contained within each. For this reason, in order to preserve the relation between systems, it is necessary to maintain the information about their order.\\
This gives rise to the segmentation task for extracting each one of them from each of the music sheets. Two different approaches were taken. Firstly, a segmentation method based on classical methods was developed and tested on the Franz Liszt scenario. Secondly, deep learning was tried on Sonatinas. It is important to highlight the difficulty that this labor represents due to the lack of standardization between scores. These may have a variable number of systems, uneven spacing, different graphic styles, etc.
Intrinsic structure of the two-handed systems was the only consistent consideration.


\subsubsection{System Segmentation: Threshold method}
\label{sec:segmentation_threshold}
\begin{figure}[t] 
    \begin{minipage}[b]{.48\linewidth}
      \centering
      \centerline{\includegraphics[width=3.5cm]{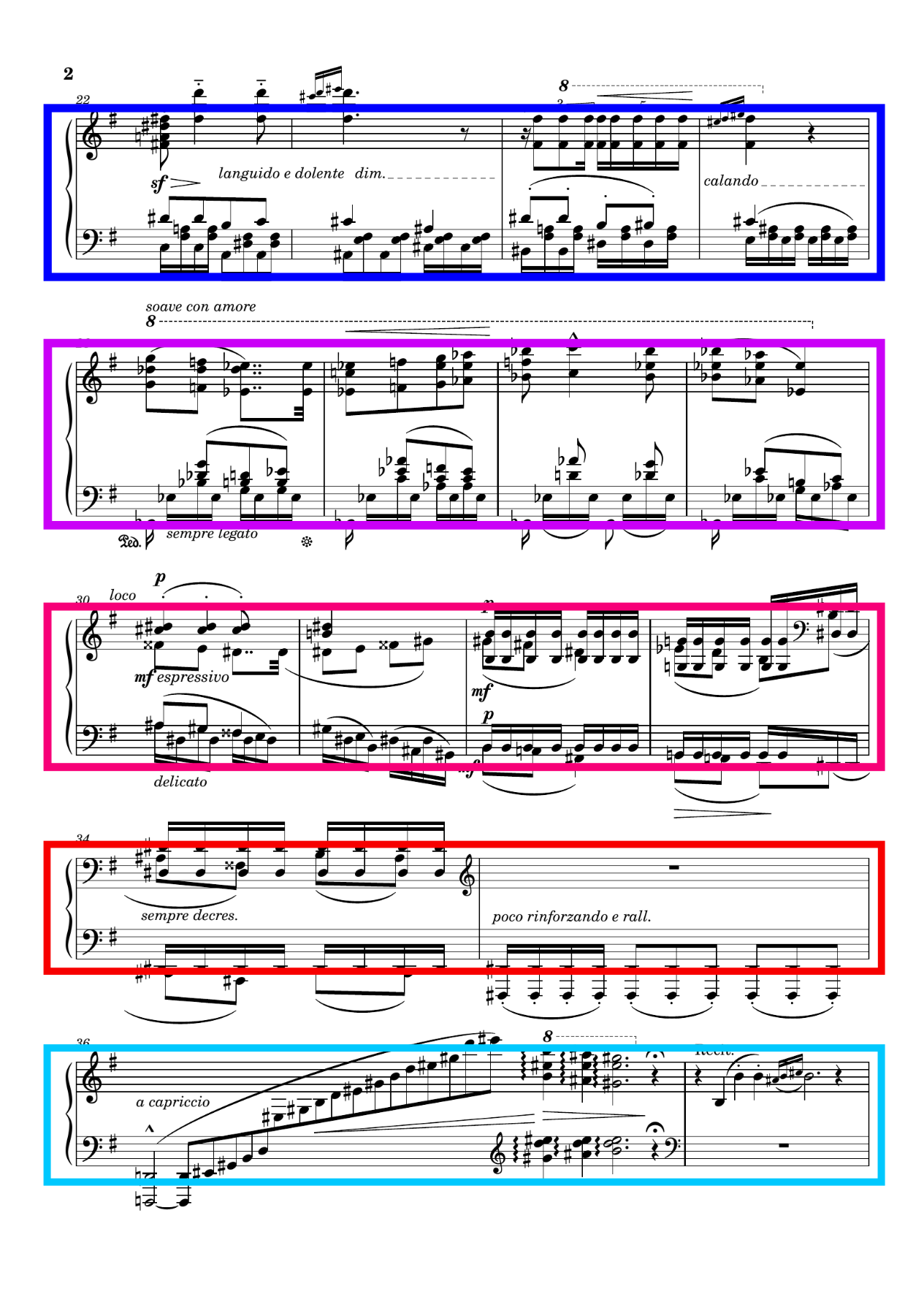}}
      \centerline{(a)}\medskip
    \end{minipage}
    \hfill
    \begin{minipage}[b]{0.48\linewidth}
      \centering
      \centerline{\includegraphics[width=2.58cm]{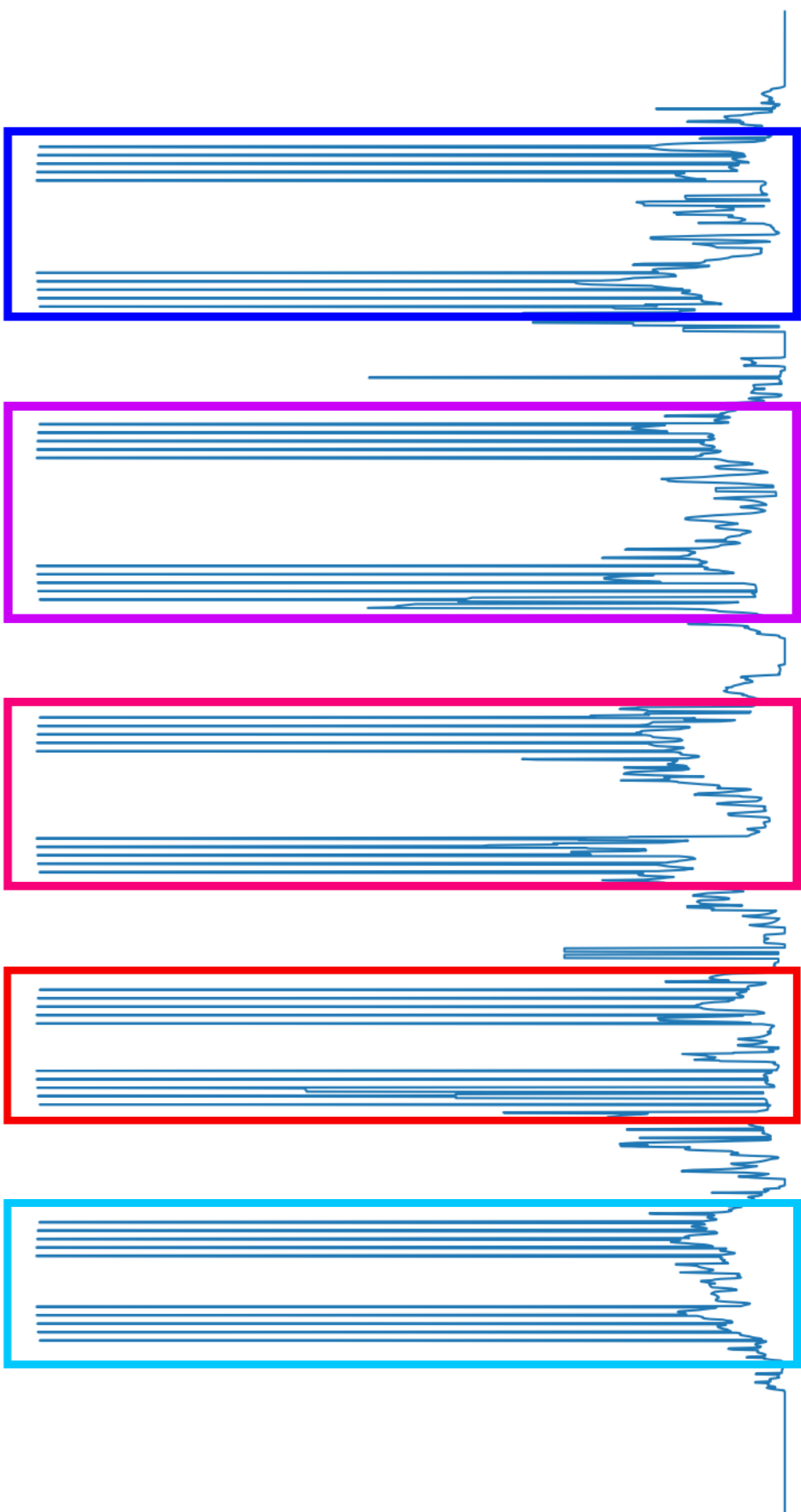}}
      \centerline{(b)}\medskip
    \end{minipage}
    \caption{(a) Example of music sheet with its systems marked. (b) Sum of the image values by rows, where the patterns of dense regions corresponding to each systems are highlighted.}
    \label{fig:score_img_sum}
\end{figure}

\begin{figure}[t] 
    \begin{minipage}[b]{.31\linewidth}
      \centering
      \centerline{\includegraphics[width=2.66cm]{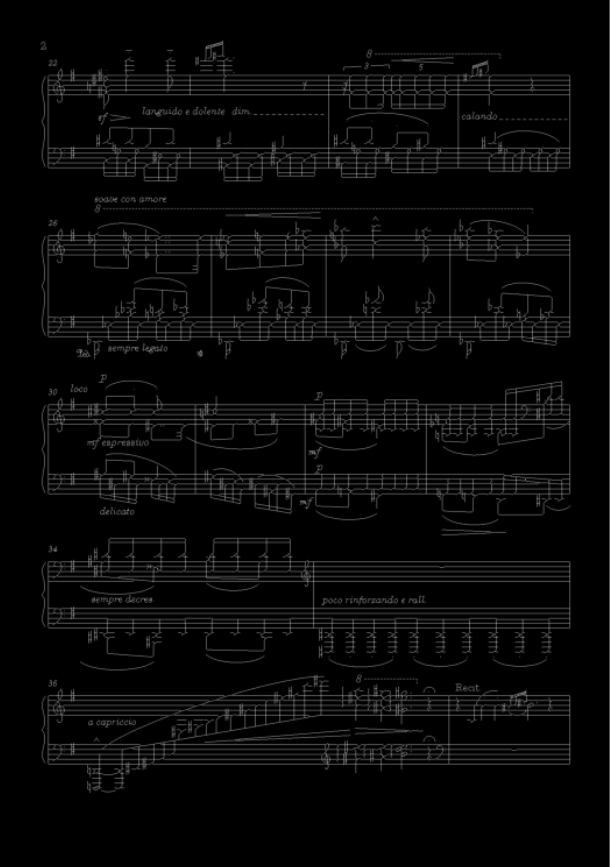}}
      \centerline{(a)}\medskip
    \end{minipage}
    \hfill
    \begin{minipage}[b]{0.31\linewidth}
      \centering
      \centerline{\includegraphics[width=2.65cm]{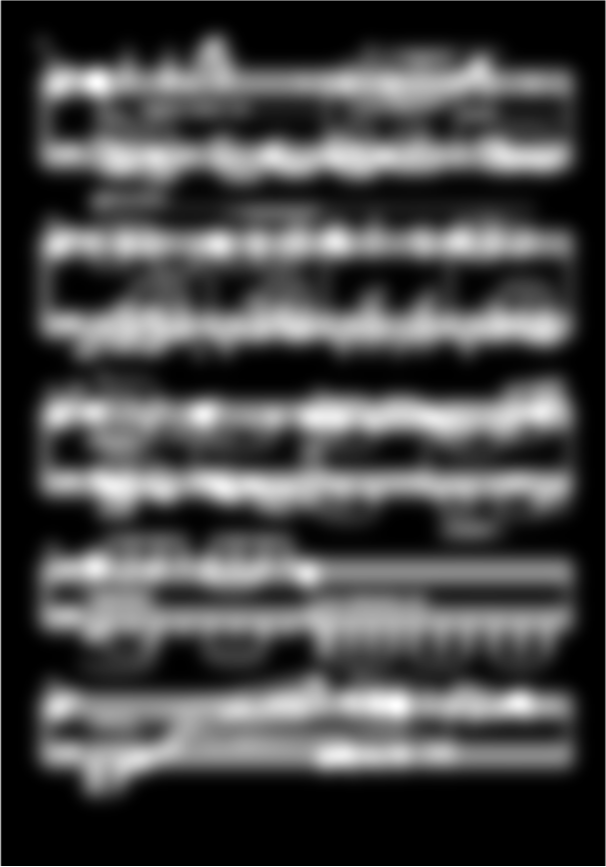}}
      \centerline{(b)}\medskip
    \end{minipage}
    \hfill
    \begin{minipage}[b]{0.31\linewidth}
      \centering
      \centerline{\includegraphics[width=2.63cm]{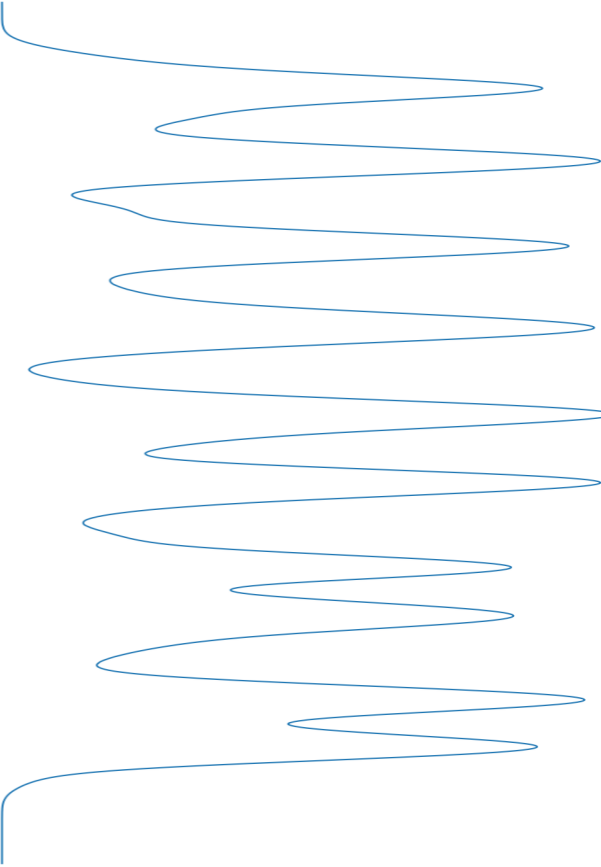}}
      \centerline{(c)}\medskip 
    \end{minipage}
    \caption{Example of different phases for segmentation with threshold method on the musical sheet shown in figure \ref{fig:score_img_sum}. (a) Score image inversion, (b) Gaussian filter and (c) the obtained profile by the sum of the image after the application of the filters.}
    \label{fig:img_filters}
\end{figure}

The first segmentation approach was developed to take advantage of the inherent characteristics of the score images, where the stave regions tend to have a greater concentration of visual elements, combined with the two-handed structure. These particularities allow us to deduce that, ideally, the systems to segment should be represented by 2 nearby and graphically dense regions, followed by space with less concentration of notes and other elements. In order to see such behavior, the sum of the image values by rows was executed. An example of the profile resulting from the sum is present in figure \ref{fig:score_img_sum} along with its corresponding music sheet.\\
Ideally, systems could be segmented using either the maxima between them, corresponding to the white spaces, or the minima, representing the staves. However, real profiles exhibit abrupt variations and noise from musical symbols. To address this, the summation curve was smoothed by applying a Gaussian filter to the image. Prior to this, image intensity inversion and skeletonization were performed to further reduce noise introduced by various musical elements. Figure \ref{fig:img_filters} shows the results of these operations, as well as the new sum profile.\\
It is observed that applying these filters smooths the curve, enabling the finite difference calculation of the derivative to identify critical points corresponding to the inter-system regions. These points define the boundaries between areas associated with each system.\\
A new challenge arises as the regions between the left and right-hand staves of the same system also correspond to minima in the curve. This requires a distinct selection criterion. To develop this method, it was assumed that intra-system regions contain a higher concentration of musical symbols than inter-system regions. Consequently, the former should correspond to minima with generally higher values than the latter within the set of minima. This translates the selection condition to check for each minimum to be lower than $a_{min}\times \mu_{min}$, where $\mu_{min}$ is the average of minimums and $a_{min}$ a multiplicative constant to adjust the value for improving the selection performance. The smoothing process primarily addresses noise; however, in certain samples, some artifacts persist, resulting in multiple closely spaced critical points. To handle these cases, an analogous approach to that used for minima was applied to select maxima ($a_{max}\times \mu_{max}$), which were then used to define the minimum grouping areas, where the lowest value was selected.\\
By means of experimentation, it was found that the optimal values for $a_{min}, a_{max}$ were $0.8$ and $0.83$ respectively. The \textit{threshold method} name comes from the fact that the selection process of critical points can be synthesized as selecting those lower than a threshold.


\subsubsection{System segmentation: Neural networks}
\label{sec:segmentation_neural}
\begin{figure}[t]
    \begin{minipage}[b]{.48\linewidth}
      \centering
      \centerline{\includegraphics[width=3.5cm]{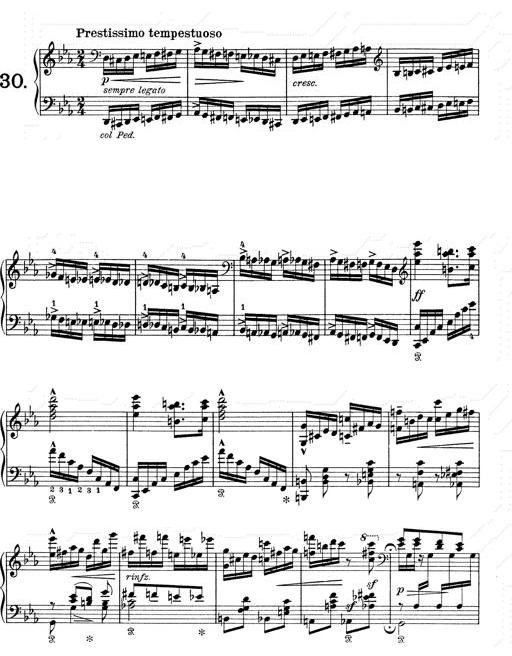}}
      \centerline{(a)}\medskip 
    \end{minipage}
    \hfill
    \begin{minipage}[b]{0.48\linewidth}
      \centering
      \centerline{\includegraphics[width=3.5cm]{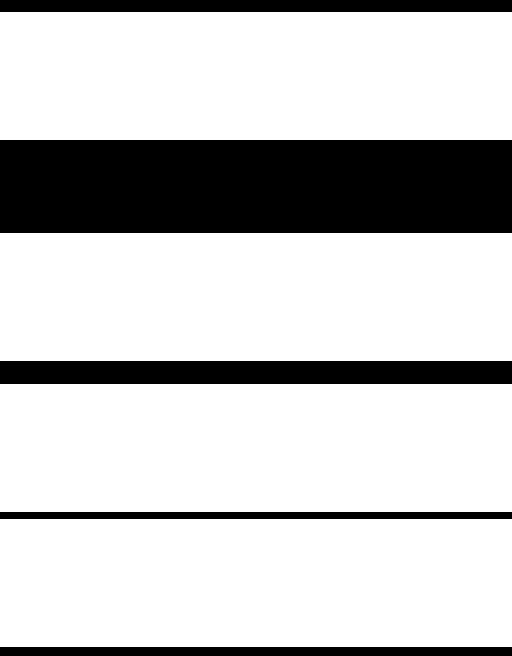}}
      \centerline{(b)}\medskip 
    \end{minipage}
    \caption{Example from the synthetic dataset randomly constructed to train segmentation neural networks. (a) Artificial sheet music image and its (b) corresponding segmentation map.}
    \label{fig:segmentation_dataset}
\end{figure}

\begin{figure}[t]
    \begin{minipage}[b]{1.0\linewidth}
        \begin{minipage}[b]{.48\linewidth}
            \centering
            \centerline{\includegraphics[width=4.0cm]{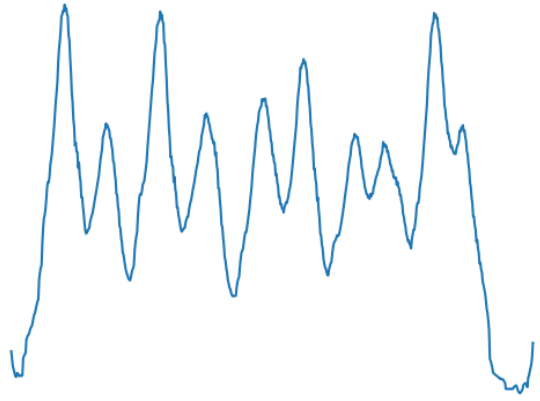}}
            \centerline{(a)}\medskip
        \end{minipage}
        \begin{minipage}[b]{0.48\linewidth}
          \centering
          \centerline{\includegraphics[width=4.0cm]{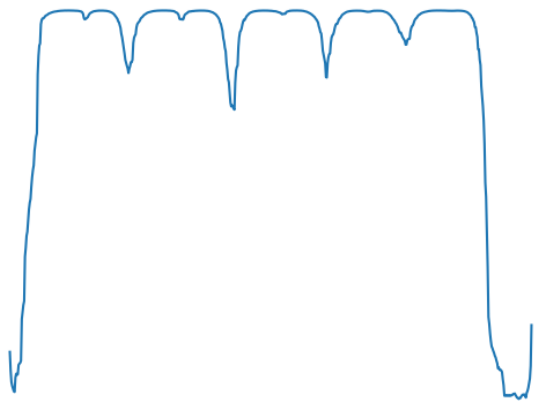}}
          \centerline{(b)}\medskip
        \end{minipage}
    \end{minipage}
    \hfill
    \begin{minipage}[b]{1.0\linewidth}
        \begin{minipage}[b]{.48\linewidth}
            \centering
             \centerline{\includegraphics[width=4.0cm]{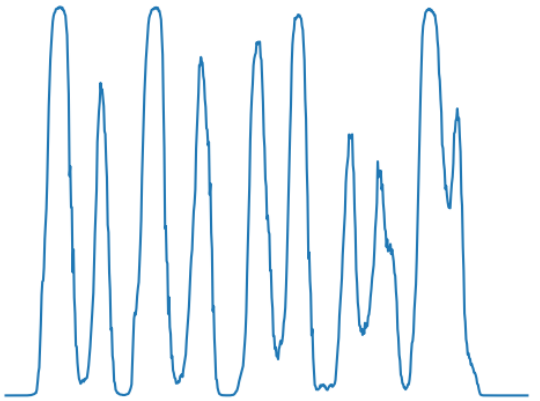}}
            \centerline{(c)}\medskip
        \end{minipage}
        \begin{minipage}[b]{0.48\linewidth}
          \centering
          \centerline{\includegraphics[width=4.0cm]{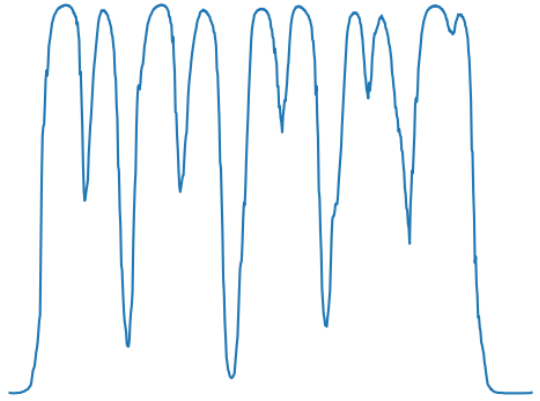}}
          \centerline{(d)}\medskip
        \end{minipage}
    \end{minipage}
    \caption{Comparison of profiles derived from the U-Net output under different processing steps: (a) raw output, (b) after sigmoid application, where profile are flattened and segmentation cues lost, (c) after subtracting a fixed value, and (d) after Cutnet’s adaptive subtraction phase, highlighting the impact of the subtraction strategy on segmentation quality by the obtained profile.}
    \label{fig:profiles}
\end{figure}

As a second segmentation approach, neural networks were considered to enhance the performance of the previous method. Particularly, the use of \textit{U-Net} \cite{unet}, as it has demonstrated its effectiveness in this task. 
The challenge lay in the absence of page segmentation maps, with only the individual Franz Liszt scenario systems available. To overcome this issue, constructing a synthetic segmentation dataset was required. This involved selecting a random number of systems from the same work, irrespective of their order, and arranging them randomly to generate artificial sheet music. Their corresponding segmentation maps were then created based on the assigned system positions.
Figure \ref{fig:segmentation_dataset} presents an example of an artificial score sheet along with its segmentation map.\\
It is evident that the segmentation map is coarse compared to those in datasets specifically designed for this task. For instance, DeepScores provides precise and well-defined maps. However, the purpose of these maps is to aid in refining the preliminary segmentation for dataset construction, rather than serving as a dedicated segmentation resource. Although beneficial, the semi-detailed construction of an instance segmentation map for each system could constitute a separate line of research.\\
Already with the set of maps, it was possible to train the U-Net model. 
The only modification made in the implementation of this architecture, with respect to its original specifications, was the number of levels, which was reduced to 3, each with a number of channels of $8,16 \text{ and } 32$, respectively.\\
When computing the row-wise sum directly from the U-Net output, the distinction between minimum points corresponding to cut marks and those that do not remains separable through thresholding. Consequently, a segmentation method analogous to the initial approach can be applied using this curve. However, when the sum is computed after applying the sigmoid function to the U-Net output, the resulting profile becomes excessively flattened. While such flattening would be beneficial if it preserved the segmentation-relevant minimum points, it instead leads to the loss of most segmentation markers.\\
To handle the observed over-flattening effect introduced by the sigmoid function, a corrective approach was tested by subtracting a constant value from the U-Net output prior to applying the sigmoid function. The modification of the profile is highly dependent on the subtracted value. If the magnitude of the subtracted value is too low, the modification will be insufficient; conversely, if it is too high, it will degrade non-cutoff minima to the extent of equating them with actual cutoff points, as illustrated in figure \ref{fig:profiles}.\\
To determine a suitable value, and given that this magnitude depends on the output values of the U-Net, a more effective approach was proposed: designing a network capable of adaptively determining this value. However, the``optimal'' subtraction values are unknown, which would have rendered this approach unfeasible. Nevertheless, the desired profile is well defined, as it corresponds to the one derived from the synthetic segmentation map, resembling a step function, taking a value of 1 in regions containing systems and 0 elsewhere. Consequently, this was established as the model's objective. To achieve this, a two-step execution model was proposed. First, the subtraction of the adaptive value. Second, the modification of the profile to match the one obtained from the segmentation map, formulated as a classification task for the points along the profile. This execution, as well as the model architecture, is described as follows:
\begin{gather}
    y(x) = \sigma ( x - ReLU( \sum_h(x)^\top W_1 + b_1) ), \\
    z(y) = \sigma \left ( U_{1:3} \circledcirc  V_{5} \circledast \left [ \tanh( V_{1:4} \circledast \sum_h(y)^{\top} ) W_2 + b_2 \right ] \right ),
\end{gather}
where $x$ represents the U-Net output, $\sum_{h}$ denotes the sum per row of the two-dimensional input, $\sigma$ is the sigmoid function, $W_1, W_2, b_1, b_2$ are the weights of the linear layers and their respective biases, $\circledast, V_{1:5}$ represent the convolution operation and its weights, and analogously, $\circledcirc, U_{1:3}$ denotes the transposed convolution operation and its weights. 
This network was named \textit{Cutnet} because its purpose was to clarify the segmentation (``\textit{cut}'') points of the sheet scores, and subsequently, to reapply the analysis of minima to identify the systems in the image.


\section{Results}
\label{sec:results}
Thus far, the imposed conditions for the dataset, as well as the data collection and segmentation processes, have been discussed. The results of each of these phases, as applied to the two defined scenarios, are now presented, ultimately leading to the construction of the dataset.

\subsection{Franz Liszt scenario}
\label{sec:results_franz}
For this scenario, $887$ PDF files were downloaded, of which $356$ were selected and processed, resulting in $5501$ page images. From these, $4013$ were chosen for segmentation. As the first constructed case, the only available method was the developed threshold-based approach, which was consequently applied to the selected images, generating $21348$ segmented samples. Due to the method's inherent limitations, a verification step was needed to ensure the segmented images were complete and free from issues such as incomplete staves or partial system cuts. After manual validation $10810$ valid systems were retained.

\subsection{Sonatinas scenario}
\label{sec:results_sonatinas}
In the case of Sonatinas, a total of $616$ PDF files were downloaded, from which $366$ were selected and processed, resulting in $5900$ page images. Of these, $3927$ were chosen for segmentation. Since the previous scenario enabled the construction of the artificial segmentation dataset, the previously discussed neural network-based methods, \textit{U-Net} and \textit{Cutnet}, were available in addition to the baseline threshold method. 
To evaluate the segmentation performance of these neural network models, standard segmentation metrics were computed over a synthetic validation dataset. These results are summarized in table \ref{tab:segmentation_metrics}. 
\begin{table}[t]
    \centering
    \begin{tabular}{|c|c|c|c|c|}
    \hline
                    & \textit{IoU} & \textit{F1} & \textit{Precision} & \textit{Recall} \\
    \hline
    \textit{U-net}  & 0.584        & 0.672       & 0.729              & 0.787           \\
    \textit{Cutnet} & 0.485        & 0.604       & 0.744              & 0.612           \\
    \hline
    \end{tabular}
    \caption{Segmentation metrics for \textit{U-net} and \textit{Cutnet} models using a synthetic evaluation dataset.}
    \label{tab:segmentation_metrics}
\end{table}
While the U-Net model achieved superior segmentation metrics, it was ultimately unsuitable for the dataset construction process. This limitation arises because using the segmentation maps directly often produced incomplete or fragmented system samples, contradicting the dataset’s requirement for structural completeness. Consequently, an analogous profile-based segmentation approach was necessary. Under this criterion, the \textit{Cutnet} model demonstrated the most reliable practical performance. Ultimately, the $3927$ images were processed, yielding $20490$ Sonatina systems, of which $14000$ were selected for inclusion.\\
Finally, to standardize the image dimensions, all samples underwent a resizing process. The resolution was $128 \times 512$ chosen to ensure compatibility with deep learning applications and preserve details. This process ultimately formed both scenarios and the complete dataset, which contains $24810$ samples. Additionally, each system has its corresponding metadata, which comprises: piece title, author, key (if available), IMSLP page, and system number. Unfortunately, the last characteristic does not always correspond to the actual system number within the piece, due to the discarded erroneous segmented images. Therefore, the number corresponds to the position of the system in relation to the set of systems belonging to the same piece. This limitation will be addressed in future updates as segmentation methods improve, with the goal of providing accurate system order to enhance the dataset’s utility for sequence-based tasks.
\section{Conclusions}
\label{sec:conclusions}
The construction of \datasetName{}, the first system-segmented image score dataset, marks a significant advancement in bridging symbolic music representation and machine learning applications. Focused on two-handed piano compositions, the dataset ensures structural consistency and format uniformity, offering $24.8$k high-quality, standardized samples with detailed metadata and system-level segmentation. \\
By addressing key limitations of existing datasets, \datasetName{} provides a structurally consistent and information-rich resource suitable for applications beyond traditional Optical Music Recognition (OMR). Its standardized image format captures detailed compositional content, timing, and system-level order, making it ideal for tasks involving sequential pattern recognition, hierarchical modeling, and symbolic music interpretation. This structural clarity also extends its applicability to layout analysis, structured data modeling, and cross-modal tasks integrating music with text, audio, or visual content.\\
Future work will focus on enhancing segmentation precision, refining metadata completeness—particularly system ordering—and expanding the dataset to include additional composers, composition types, and a greater number of samples. These improvements aim to further increase the dataset’s versatility and impact across music-related and machine learning tasks.\\
In summary, \datasetName{} introduces a novel and well-structured dataset that connects symbolic music representation with machine learning by combining high structural integrity with detailed metadata. Its design supports broad research in both music and machine learning, offering a strong foundation for future developments.

\newpage
\bibliographystyle{IEEEbib}
\bibliography{refs}

\end{document}